\documentclass[a4paper,conference]{IEEEtran}
\ifCLASSINFOpdf
\else
\fi

\ifCLASSINFOpdf
\else
   \usepackage[dvips]{graphicx}
\fi
\usepackage{url}
\usepackage{graphicx}
\usepackage{amsmath}
\usepackage{amsfonts} 
\usepackage{amssymb}

\begin{document}

%
\title{Video Captioning in Compressed Video}


\author{
	Mingjian Zhu$^{1,2}$, Chenrui Duan$^{1,2}$, Changbin Yu$^{1,2}$\\
		\normalsize $^1$Zhejiang University, China\\
			\normalsize $^2$School of Engineering, Westlake University, China \\
			\normalsize \{zhumingjian, duanchenrui, yu\_lab\}@westlake.edu.cn\\		
	}

\maketitle

%


%




\begin{abstract}
Existing approaches in video captioning concentrate on exploring global frame features in the uncompressed videos, while the free of charge and critical saliency information already encoded in the compressed videos is generally neglected. We propose a video captioning method which operates directly on the stored compressed videos. To learn a discriminative visual representation for video captioning, we design a residuals-assisted encoder (RAE), which spots regions of interest in I-frames under the assistance of the residuals frames. First, we obtain the spatial attention weights by extracting features of residuals as the saliency value of each location in I-frame and design a spatial attention module to refine the attention weights. We further propose a temporal gate module to determine how much the attended features contribute to the caption generation, which enables the model to resist the disturbance of some noisy signals in the compressed videos. Finally, Long Short-Term Memory is utilized to decode the visual representations into descriptions. We evaluate our method on two benchmark datasets and demonstrate the effectiveness of our approach.
\end{abstract}


%

\section{Introduction}

Video captioning, which aims to automatically describe video content in natural language, becomes progressively popular in the communities of computer vision and natural language. Earlier works on video captioning~\cite{venugopalan2014translating,xu2017learning, wang2018reconstruction} mostly adopt the typical deep encoder-decoder framework and achieve encouraging performance. In the encoding stage, the features of sampled frames are extracted by CNN, when RNN is employed to encode the sequences of features into visual representations. Then a decoder is utilized to translate the representations into descriptive sentences. But such a typical framework basically suffers from two severe drawbacks. First, a pretrained CNN is generally employed to directly extract the features of the global RGB images, which neglects that spatial distribution of significant signals in a single frame is generally imbalanced and lacks explicitly procedure of highlighting the meaningful signals in the salient regions. To tackle this problem, some previous works~\cite{wang2018spotting, tu2017video} propose region level attention mechanisms. However, these methods generate spatial attention maps mainly by exploring the information from the RGB image itself. An efficient method with the capability of simultaneously exploiting the information from RGB images and incorporating other visual sources is believed to achieve better performance and is eagerly needed. Second, an obvious fact is that a video can be compressed to a fairly small size, which implies that an uncompressed video contains high information redundancy. The boring and repeating patterns of decoded frames in videos will largely drown the interesting signals and hinder the exploration of essential information. These are common problems existing in video understanding tasks.
\begin{figure}[!t]
\includegraphics[width=1.0\linewidth]{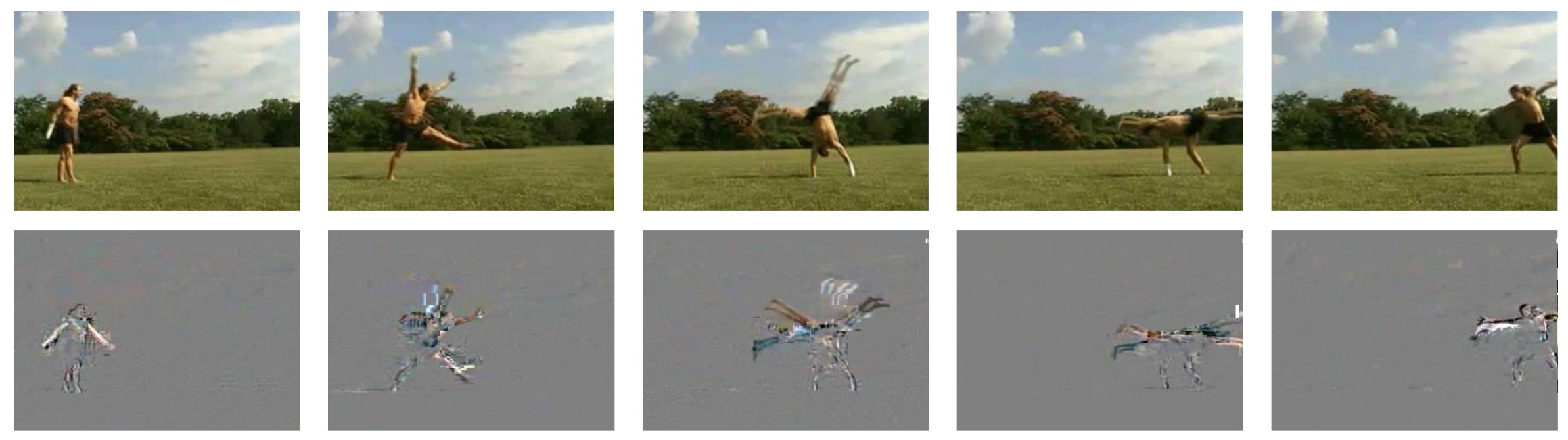}
\caption{Example I-frames (top) and residuals (bottom) shown in two rows. The residuals frames contain lots of noisy movement patterns, which correspond to the regions of interest in I-frame. (Best viewed in color)}
\label{fig:motivation}
\end{figure}

To address these issues, recent works~\cite{wu2018compressed, shou2019dmc, wang2019fast} apply deep models directly on compressed videos, which is encoded by the compression standards like MPEG-4, H.264, and HEVC. Tackling video understanding problem in compressed videos brings lots of benefits. First, the saliency information is already embedded in the compressed videos. Accessing inherent information costs little computation. Second, the compression technique reduces excess information in a video and keeps the most essential signals, i.e., the saliency signals. The saliency signals leave out the visual appearance variations and highlight the prominent information.
Generally, an encoded video is comprised of a number of GOPs (group of pictures) and the GOP contains two kinds of frames: I-frames and P-frames~\cite{wu2018compressed, shou2019dmc, wang2019fast}. I-frame in a compressed video can be regarded as a regular complete image. P-frames encode only the 'change' among the frames. P-frames can be reconstructed by residuals and motion vector~\cite{wang2019fast}. As the example provided in Fig.~\ref{fig:motivation}, The residuals, though noisy, resemble optical flow~\cite{sun2018optical} and locate the areas with visually drastic changes in the video. The salient regions in the residuals frames exhibit notable motion patterns, which correspond to the actions or objects in the videos and human descriptions. In the video captioning task, the captions for a video are labeled by humans, and human attention tends to be attracted by the salient regions where the visual appearance changes noticeably. Following this observation, we propose to explore and leverage the saliency information embedded in the compressed videos to guide the generation of spatial attention.

\begin{figure*}[!htb]
\centering
\includegraphics[width=1.0\linewidth]{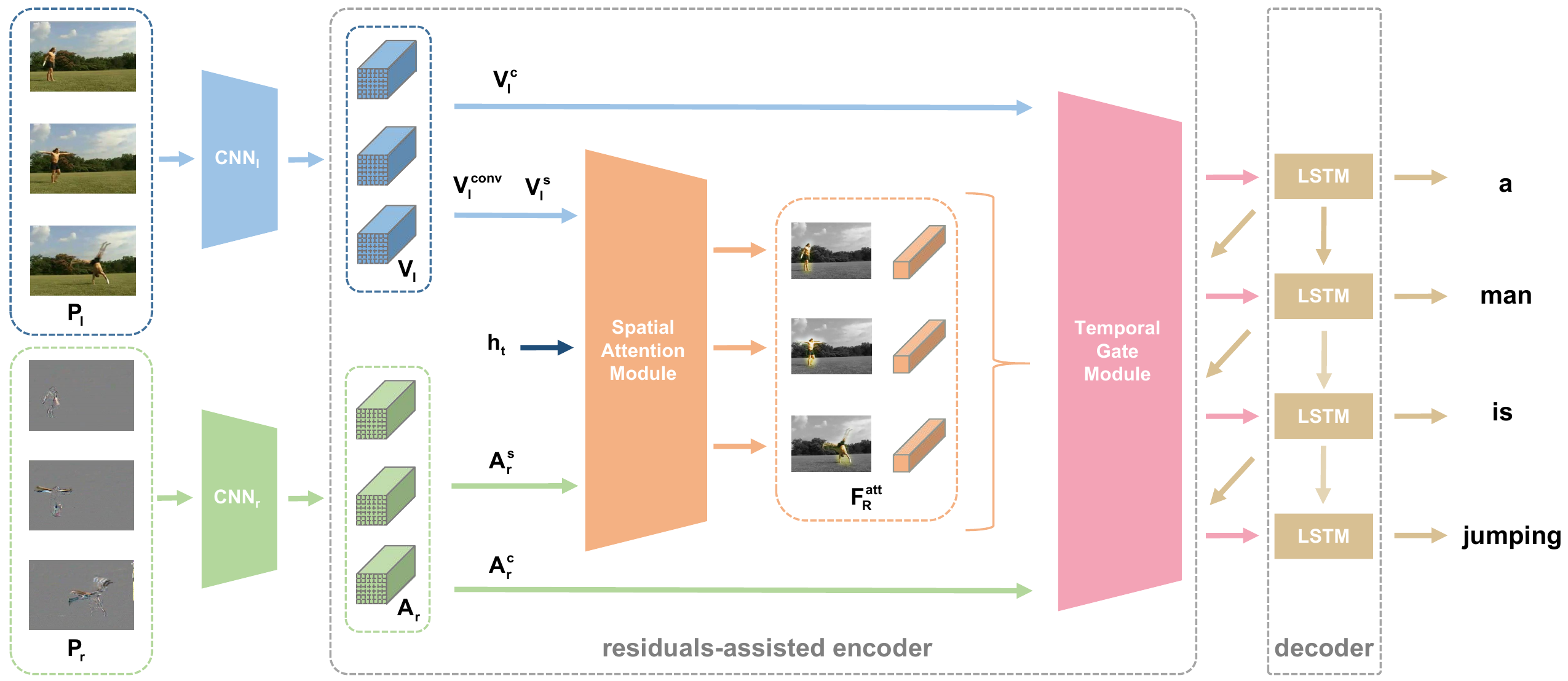}

\caption{Illustration of our video captioning framework. It contains two components: (1) residuals-assisted encoder (2) decoder. Specifically, residuals-assisted encoder contains the spatial attention module and temporal gate module, and automatically selects the most correlated visual features as the input to the decoder. The decoder, i.e., LSTM in this framework, is utilized to generate the description word by word.
(Best viewed in color)}

\label{fig:framework}
\end{figure*}

In this paper, we propose a novel video captioning method, called video captioning in compressed video (VCCV). This method generates the spatial attention map for the I-frame, with the assistance of saliency information in the residuals. We also design a gate mechanism to adaptively fuse visual features for description generation with the motivation of retaining only the positive impact of the residuals. Our proposed method takes advantage of signals from different sources(i.e. I-frame and residuals). The main contributions of our work can be summarized as follows: 

\begin{itemize}
\item We propose an end-to-end deep framework to efficiently exploit I-frame and residuals in the compressed videos. As far as we know, we are the first to tackle video captioning task with the usage of the compressed signals. Besides, comparing with previous video understanding methods on compressed videos, we exploit the saliency information in residuals frame to assist the generation of spatial attention map in the deep model.

\end{itemize}

\begin{itemize}
\item  Our residuals-assisted encoder (RAE) attends salient regions in I-frame with the assistance of the residuals. Besides, the gate mechanism in RAE fuses visual features and inputs them to the decoder, which enhances the robustness of the method against the noise in the residuals. 
\end{itemize}
 
\begin{itemize}
\item Our method is evaluated on two public benchmarks: MST-VTT~\cite{xu2016msr} and Charades~\cite{sigurdsson2016hollywood} and achieves results that rival state-of-the-art methods.
\end{itemize}

\section{Related Work}

\subsection{Video Captioning}
\noindent \textbf{Template Based Video Captioning.} Inspired by the progress in image captioning, some earlier works employ the template to synthesize the sentence in video captioning task. Template based language model can ensure the grammatical correctness.~\cite{kojima2002natural} correlates concept hierarchy of actions with semantic features of human motions for video captioning task.~\cite{rohrbach2013translating} proposes to employ conditional random fields and a template model to generate a sentence for a video. Although template based methods can ensure the completeness of sentences, their generated descriptions are inflexible.
\\
\\
\noindent \textbf{Encoder-Decoder Based Video Captioning.} 
Mean Pool method~\cite{venugopalan2014translating} extracts the features of each frame by using CNN, mean pool the features across the entire video, and input this to the LSTM network. a-LSTM~\cite{gao2017video} introduces semantic cross-view loss and relevance loss to enforce the consistency between the features of video and language. Attention Fusion ~\cite{hori2017attention} propose a  method to selectively attend specific modalities of features, which provides a way to fuse multimodal information for video captioning. 
RecNet~\cite{wang2018reconstruction} contains three parts: the CNN-based encoder, the LSTM-based decoder, and the reconstructor. The proposed two types of reconstructors aim to minimize the diversity between the original and reproduced video features.  \cite{wang2018reconstruction} also proposes to exploit both video-to-sequence and sequences-to-video flows for video captioning. SAM~\cite{wang2018spotting} designs a model to spot and aggregate the salient regions in the video. This method explores the information from the RGB image to locate the salient regions, while our VCCV exploits the residuals frames in the compressed video.~\cite{zhu2019attention} proposes Attention-based Densely Connected Long Short-Term Memory which makes all previous cells connected to the current cell and the updating of the current state related to all its previous states.

\subsection{Video Understanding Model on Compressed Videos}
~\cite{zhang2018video, khatoonabadi2017compressed} propose methods for saliency estimation in compressed videos, without using deep models. Only a few previous works apply deep learning directly on compressed videos and neither of them tackles video captioning task. For recognizing the actions in the compressed videos, CoViAR~\cite{wu2018compressed} develops three independent deep CNN models, which are trained on data of different modalities: I-frame, residuals, and motion vector. Their outputs are fused together as the final prediction. However, the correlation among the three modalities cannot be fully exploited because of the model independence. We argue that the relationships among different modalities should be further explored to generate more discriminative visual representation. In order to achieve state-of-art performance, CoViAR employs optical flow feature, which consumes lots of computation. To avoid optical flow computation in the inference stage, DMC-Net~\cite{shou2019dmc} leverages Generative Adversarial Networks (GAN) to approximate optical flow with the usage of residuals and motion vector. However, optical flow is still needed in training. Motion aided Memory
Network (MMNet)~\cite{wang2019fast} utilizes residuals as input and leverages motion vector to align the hidden features. The generated cell features of the modified LSTM are aggregated to the detection network for the video object detection task. MMNet reveals that special patterns exist in the features generated by compressed signals. This finding inspires us to explore the potential of fully exploiting the patterns to contribute to the visual encoding stage in video captioning task. However, according to the above three works, the signals in compressed videos are noisy, which hinders their direct application as features for video encoding. To address this issue, we propose a more practical method, which generates attended spatial representation in the deep model by exploiting the embedded saliency information in the compressed videos.

\section{METHOD}
As shown in Fig.~\ref{fig:framework}, the proposed VCCV method is comprised of two components, namely the residuals-assisted encoder and the decoder. The residuals-assisted encoder is designed to encode the visual content information, while the decoder is used for generating sentences. Different from many previous works, which explore the relationship between visual and context information in video-sentences pairs, we focus on designing a video captioning model that effectively generates discriminative visual representations. 

A compressed video contains I-frames and residuals. $N$ I-frames are sampled uniformly in the compressed video. We also collect the residuals frame in the corresponding GOP of each I-frame.  The shapes of the sampled I-frames $P_I$ and residuals $P_r$ are $(N, H_P, W_P, C)$, where $H_P$, $W_P$, and $C$ denote the height, width, and the channels of the frames, respectively. We leverage two pretrained CNNs to extract the features of I-frames and their corresponding rough attention maps from residuals: 
\begin{equation}
\begin{aligned}
V_I = CNN_{I}(P_I),\\ A_{r} = CNN_{r}(P_{r}),
\end{aligned}
\end{equation}
where $ V_I \in \mathbb{R}^{ N\times H \times W \times D_I}$ is the extracted feature of I-frame, and $A_r \in \mathbb{R}^{N \times H \times W \times D_r}$ is the feature of the residuals. Since the salient regions in residuals frames correspond to notable actions or objects in I-frames, $A_r$ can be considered as the rough attention map. $D_I$ and $D_r$ are the output channels of pretrained backbone model $CNN_I$ and $CNN_r$, respectively. Note that we extract and store $V_I$ and $A_r$ before training our proposed model.

\subsection{Residuals-Assisted Encoder}
\label{section:Residuals-Assisted Encoder}
After the pre-processing procedure, the resulting features $V_I$ and $A_r$ are fed into residuals-assisted encoder to generate attended visual representation $F_G^{att}$:
\begin{equation}
\begin{aligned}
F_G^{att} &= RAE(V_I, A_r, h_{t-1})
\end{aligned}
\end{equation}
where $h_{t-1}$ denotes the hidden state of decoder in time step $t-1$. We conduct the average pooling operations along the spatial dimensions to get the following features:
\begin{equation}
\begin{aligned}
V_{I}^{c} &= \frac{1}{H \times W} \sum_{h=1}^{H}\sum_{w=1}^{W} V_I, \\
\end{aligned}
\end{equation}
\begin{equation}
\begin{aligned}
 A_{r}^{c} &= \frac{1}{H \times W}\sum_{h=1}^{H}\sum_{w=1}^{W} A_{r},\\
\end{aligned}
\end{equation}
We also conduct the average pooling operations along the channel dimensions:
\begin{equation}
\begin{aligned}\label{channel pooling}
V_{I}^{conv} &= Conv(V_I), \\
\end{aligned}
\end{equation}
\begin{equation}
\begin{aligned}
V_{I}^{s} &= \frac{1}{D_r} \sum_{d=1}^{D_r} V_{I}^{conv},\\
\end{aligned}
\end{equation}
\begin{equation}
\begin{aligned}
A_{r}^{s} &= \frac{1}{D_r} \sum_{d=1}^{D_r} A_r,\\
\end{aligned}
\end{equation}
where the convolution operation in Eq.\ref{channel pooling} reduces the channel dimension of $V$ from $D_I$ to $D_r$ and we train this layer along with the encoder-decoder model in an end-to-end manner. $V_I^s \in \mathbb{R}^{N \times H \times W}$, $A_r^s \in \mathbb{R}^{N \times H \times W}$ and $V_I^{conv} \in \mathbb{R}^{N \times H \times W \times D_r}$ are input to spatial attention module (SAM), while $V_I^c \in \mathbb{R}^{N \times D_I}$ and $A_r^c \in \mathbb{R}^{N \times D_r}$ are fed into temporal gate module (TGM).\\
\textbf{Spatial Attention Module.}  The SAM refines the rough attention map by incorporating the spatial information from I-frame and context information from decoder. For each I-frame and its corresponding residuals frame in the same GOP, SAM produces the refined attention map $\alpha_{R}$ by using $h_{t-1}$, $V_I^s$ and $A_r^s$:
\begin{equation}
\begin{aligned}
\alpha_{R} &= \tanh(\mathbb{E}_{\alpha} (W_{t} h_{t-1}) + W_{I} V_I^{s} + W_{r} A_{r}^{s} )\label{eq:alpha_R}
\end{aligned}
\end{equation}
where $W_{t}, W_{I}, W_{r}$ are learnable weights. $\mathbb{E}_{\alpha} (\cdot)$ operation expands the size of tensor from  $H \times W$ to $N \times H \times W$.
The attention weights of each region are computed by applying softmax function along the spatial dimensions:
\begin{equation}
\begin{aligned}
A_{R} &=\frac{exp(\alpha_{R})}{\sum_{h=1}^{H}\sum_{w=1}^{W} exp(\alpha_{R})}
\end{aligned}
\end{equation}
After that, we obtain the attended global feature $F_{R}^{att}$ by a weighted average over the regional features:
\begin{equation}
\begin{aligned}
F_{R}^{att} &=  \frac{1}{H \times W}\sum^{H}_{h=1} \sum^{W}_{w=1} \mathbb{E}_{A}({A_{R}}) \odot V_{I}^{conv},
\end{aligned}
\end{equation}
where $\mathbb{E}_{A}(\cdot)$ expands the size of $A_R$ from $N \times H \times W$ to $N \times H \times W \times D_r$. $\odot$ denotes the element-wise hadamard product.\\
\textbf{Temporal Gate Module.} Although the attended global feature contains sufficient appearance information, it cannot eliminate the negative impact of noisy signals in the compressed videos. To further alleviate this problem, our proposed TGM fuses the original visual feature $V_{I}^{c}$ and the spatially attended feature $F_{R}^{att}$ with a generated confidence score.
Considering the context information in the hidden state of the LSTM-based text decoder, the distribution of noisy signals in residuals, and the visual information from the original extracted feature, the confidence score $G \in \mathbb{R}^{N}$ is defined as:
\begin{equation}
\begin{aligned}
G &= \tau (W_G^{T} \cdot ( W_{Gt} h_{t-1} + W_{Gr} A_{r}^{c} + W_{GI} V_I^c))
\end{aligned}
\end{equation}
where $W_G$, $ W_{Gt}$, $ W_{Gr}$ and $W_{GI}$ denote the parameters to be learned,
and $\tau$ is the sigmoid activation function. Different from temporal attention mechanism in previous work~\cite{tu2017video, song2017hierarchical}, our proposed TGM controls how much the spatial attended feature and the original extracted feature respectively contribute to the encoded visual representation, which differs in the temporal dimension. The final encoded visual representation $F_G^{att}$ is defined as:
\begin{equation}
\begin{aligned}
F_{G}^{att} &= \mathbb{E}_{G}(G) \odot W_{GR}F_{R}^{att} + \mathbb{E}_{G}(1-G) \odot W_{GI}V_I^c
\end{aligned}
\end{equation}
where $\mathbb{E}_{G}(\cdot)$ expands the sizes of $G$ and $(1-G)$ from $N$ to $N \times D_I$. Again, $\odot$ denotes the element-wise product. By taking advantage of two features, the fusion mechanism can efficiently alleviate the disturbance from some noisy residuals. 
After obtaining the visual representation $F_{G}^{att}$ in the TGM, we further perform the ordered combination of the following operations to this representation: average pooling along frames dimension, linear transformation, ReLU, and dropout.

\subsection{Decoder}
\label{section:Decoder}
Our decoder 
adopts the widely used Long Short-Term Memory (LSTM)~\cite{hochreiter1997long} and takes video representation $F_G^{att}$ from RAE as input:
\begin{equation}
\begin{aligned}
h_t, c_t = LSTM([F_G^{att};x_{t-1}], (h_{t-1}, c_{t-1})), 
\end{aligned}
\end{equation}
 where $[\cdot;\cdot]$ stands for the vector concatenation operation. $h_t$ and $c_t$ are the hidden state and cell state of the LSTM at time step $t$. $x_{t-1}$ is the texture feature of the word at time step $t-1$. After obtaining the hidden state $h_t$, a linear layer is applied after the LSTM layer and a softmax layer are leveraged to produce the probability distribution over all the vocabulary words. In training stage, the ground truth word at previous time step is utilized when predicting the current word. In testing stage, the previous predicted word is input to the LSTM. We employ beam search method~\cite{freitag2017beam} to generate sentences. The framework can be trained in an end-to-end manner and the loss function we are optimizing is the log-likelihood:
\begin{equation}
   \mathop{max}\limits_{\theta}\sum_{t=1}^T log P_r(y_t|F_G^{att}, y_{t-1};\theta),
\end{equation}
where $T$ denotes the length of the sentences. $y_t$ stands for the word of $t$-th time step and $\theta$ denotes the trainable parameters in our method.

\section{Experiments}
\subsection{Datasets, Metrics, Preprocessing and Details}
\noindent\textbf{Datasets.}
{MSR-VTT dataset}~\cite{xu2016msr} is a large-scale video description dataset for evaluating video captioning methods.
It contains 10K web video clips with 41.2 hours in 20 categories. Each video
clip is annotated with about 20 natural sentences. We divide the dataset into training, validation, and testing with 6,513 clips, 497 clips, and 2,990 clips respectively as the original split in the MSR-VTT dataset.

{Charades dataset}~\cite{sigurdsson2016hollywood}  contains 9,848 video clips with 27,847 video descriptions.  Following the official split~\cite{sigurdsson2016hollywood}, we utilize 1863 video clips for evaluation, and the other video clips are used for model development.

\begin{table*}[htb]
    \normalsize
	\centering
	\caption{Comparison of model variants on MSR-VTT dataset. (The best scores are \textbf{bold})}
	\begin{tabular}{l|c|c|c|c|c|c|c}
		\hline
		{Model} & BLEU@1 & BLEU@2 & BLEU@3 & BLEU@4 & METEOR & CIDEr & ROUGE-L\\ 
		\hline
		{I-frame} &71.1 &55.7 &41.7  &29.9 &23.5 &29.7 &55.0 \\
		\hline
        {RAE w/o gate and residuals} &78.1 &63.0 &49.1 &{37.6} &{27.5}  &{44.7} &{58.5} \\
 		\hline
		{{RAE w/o gate}}  &78.6 &64.5 &51.2 &39.6 &27.8 &\textbf{46.0} &59.8 \\
		\hline
		{{RAE(full)}}  &\textbf{78.6} &\textbf{64.8} &\textbf{51.7}   &\textbf{40.0} & \textbf{28.0} & {45.7} & \textbf{60.0}\\
		\hline
	\end{tabular}%
	\label{tab:model varient msrvtt}%

\end{table*}%

\begin{table*}[htb]
\normalsize
	\centering
	\caption{Comparison of model variants on Charades dataset. (The best scores are \textbf{bold})}

	\begin{tabular}{l|c|c|c|c|c|c|c}
		\hline
		{Model} & BLEU@1 & BLEU@2 & BLEU@3 & BLEU@4 & METEOR & CIDEr & ROUGE-L\\ 
		\hline
		{I-frame} &36.9 &27.6 &18.6 &12.3 &14.0 &8.0 &38.4\\
		\hline
        {RAE w/o gate and residuals} &{52.2} &{36.3}  &{24.2} &{16.3} &{17.5}  &{20.5} &{41.5}\\
 		\hline
		{{RAE w/o gate}}  &\textbf{54.2} &36.5 &{23.9} &15.9 &18.0 &{20.6} &41.3\\
		\hline
		{{RAE(full)}} &{53.2} & \textbf{36.6} & \textbf{24.3} & \textbf{16.5} & \textbf{18.0} & \textbf{21.0} & \textbf{41.5}\\
		\hline
	\end{tabular}%
	\label{tab:model varient charades}%
\end{table*}%

\noindent\textbf{Evaluation Metrics.} In this paper, we employ several common metrics to evaluate our proposed models: BLEU~\cite{papineni2002bleu}, METEOR ~\cite{denkowski2014meteor}, CIDEr~\cite{vedantam2015cider} and ROUGE-L~\cite{lin2004automatic}. These metrics are widely used in image/video captioning tasks. For a fair evaluation, We utilize the Microsoft COCO evaluation toolkit~\cite{chen2015microsoft} to compute all the values in this paper and report them as percentage(\%).

\noindent\textbf{Preprocessing.} 
We leverage the CNN model as a visual feature extractor and compute the descriptor with a fixed number of 20 frames. I-frame is a full image. The features of the I-frames are extracted by a CNN pretrained in Imagenet dataset~\cite{russakovsky2015imagenet}. We extract the features of I-frame in our method with backbone model ResNet-152~\cite{he2016deep} and the size of the extracted features from the penultimate layer is $2048\times7\times7$. 
Similar to~\cite{wu2018compressed}, We train a  ResNet-18 model for video action recognition by utilizing residuals frames in HMDB-51 dataset~\cite{jhuang2011large}. For feature extraction, we store the activations from the penultimate layer of this model with a size of $512\times7\times7$.

In the description preprocessing stage, a vocabulary for the corpus is built. All words in the captions are converted to lower case and the sentences are tokenized. The extremely rarely seen words in the vocabulary which appear less than 3 times are also removed.
The max length of sentences is set as 50 and the provided caption with length more than that number will be cut short. During the training phase,  we add a begin-of-sentence $<$BOS$>$ tag at the beginning of each caption, and an end-of-sentence $<$EOS$>$ tag at its end.  In the testing phase, we input the $<$BOS$>$ tag to the decoder for the first time-step, then the word is predicted one by one until the $<$EOS$>$ tag appears. Each word in the processed sentences will be converted into a one-hot vector. The unseen words in the vocabulary are set to the $<$UNK$>$ flags.

\noindent\textbf{Training Details.} We optimize our model using Adam~\cite{kingma2014adam} optimizer with an initial learning rate of $1\times10^{-4}$. In our model, the word embedding size is set as 500 and the dimension of the LSTM hidden size is set to be 512. The dimension of visual representation input to the decoder is 2048. To alleviate the overfitting problem, We apply dropout with a rate of 0.5 on the output of LSTM. The training batch size is set to 8 and the beam size in testing is 5.

\subsection{Ablation Study}
In Tab.~\ref{tab:model varient msrvtt} and Tab.~\ref{tab:model varient charades}, we perform ablation study to further investigate the contribution from each component of our proposed RAE to the whole system. 
All the model variants are based on the ResNet-152 backbone model and our proposed method achieves the best result in both MSR-VTT dataset and Charades dataset. 
In MSR-VTT dataset, compared with the full version of RAE, the scores of the proposed method without temporal gate module are decreased. We further remove the extracted feature of residuals in Eq.~\ref{eq:alpha_R} and the performance reduces in terms of all metrics, which demonstrates that the information in residuals can bring performance improvement. These results imply that our proposed SAM and TGM lead to a performance boost. The I-frame method is the simplified model of Mean Pool~\cite{venugopalan2014translating} in Tab.~\ref{tab:msr-vtt} with the second layer of the LSTM in decoder removed. We average the extracted features of 20 I-frames to obtain the visual representations, which are fed into the LSTM decoder for description generation. It can be observed that our framework performs better than I-frame method by a large margin. The results in the Charades dataset also validate the effectiveness of each component in RAE.

\subsection{Comparison of different methods}
Our method proposes that exploring the saliency information in the compressed videos benefits the generation of video descriptions.
To validate the effectiveness of this method, we compare our video captioning method with the existing methods on the MSR-VTT dataset in Tab~\ref{tab:msr-vtt} and Charades dataset in Tab.~\ref{tab:charades}. R, C, and A denote ResNet-152~\cite{he2016deep}, C3D~\cite{tran2015learning}, and audio~\cite{hershey2017cnn} features, respectively.
 As can be seen, our method achieves superior performance than previous works on two datasets. We also make a comparison with the methods that rank top-3 in the MSR-VTT Challenge and denote them as v2t\_navigator~\cite{jin2016describing}, Aalto~\cite{shetty2016frame}, and VideoLAB~\cite{ramanishka2016multimodal}. It can be seen that our method achieves the best performance across the three metrics. 




\vspace{0.1cm}
\begin{table}[!h]
\normalsize
	\renewcommand\arraystretch{1.05}
	\centering
	\caption{The performance comparison with the previous methods on MSR-VTT dataset. (The best scores are \textbf{bold})}

\begin{tabular}{l|c|c|c}
\hline
{Methods}  
& BLEU@4  & METEOR & CIDEr \\ 
\hline
v2t\_navigator~\cite{jin2016describing} &40.8 &28.2 &44.8 \\
Aalto~\cite{shetty2016frame} &39.8 &26.9 &45.7 \\
VideoLAB~\cite{ramanishka2016multimodal} &39.1 &27.7 &44.1 \\
\hline
Mean Pool~\cite{venugopalan2014translating} &30.4 &23.7 &35.0 \\

Attention Fusion~\cite{hori2017attention} &39.7 &25.5 &40.0 \\

MA-LSTM~\cite{xu2017learning} &36.5 &26.5 &41.0 \\

a-LSTM~\cite{gao2017video} &38.0 &26.1 &43.2 \\
TDDF~\cite{zhang2017task} &37.3 &27.8 &43.8 \\
PickNet~\cite{chen2018less} &41.3 &27.7 &44.1 \\
RecNet~\cite{wang2018reconstruction} &39.1 &26.6 & 42.7 \\
MCNN+MCF~\cite{wu2018multi} &38.1 &27.2 &42.1 \\
STAT~\cite{yan2019stat} &39.3 &27.1 &43.8 \\
DenseLSTM~\cite{zhu2019attention} &38.1 &26.6 &42.8 \\
TDConvED~\cite{chen2019temporal} &39.5 &27.5 &42.8 \\

\hline
{VCCV(R)} & {40.0} & {28.0} & {45.7} \\
{VCCV(R + A)}  &{42.2} & {28.8} & {47.2} \\
\textbf{VCCV(R + A + C)}  &\textbf{42.8} & \textbf{29.3} & \textbf{48.7} \\

\hline
\end{tabular}

\label{tab:msr-vtt}
\end{table}

\vspace{0.1cm}

\begin{table}[!h]
    \normalsize
	\renewcommand\arraystretch{1.05}
	\centering
	\caption{The performance comparison with the previous methods on Charades dataset. (The best scores are \textbf{bold})}

	\begin{tabular}{l|c|c|c}
		\hline
		{Methods}  
	    &BLEU@4 &METEOR &CIDEr  \\ \hline
        S2VT~\cite{Venugopalan_2015_ICCV} &11.0 &16.0 &14.0 \\
        SA~\cite{yao2015describing}  &7.6 &14.3 &18.1  \\
		MAAM~\cite{Fakoor2016Memory} &11.5 &17.6 &16.7 \\
		TSA-ED~\cite{wu2018interpretable}  &13.5 &17.8 &20.8  \\
		\hline
		{\textbf{VCCV(R)}} &\textbf{16.5} &\textbf{18.0} &\textbf{21.0}\\ 
		\hline
	\end{tabular}%

	\label{tab:charades}%
\end{table}

\begin{figure*}[!h]
\centering
\includegraphics[width=1.0\linewidth]{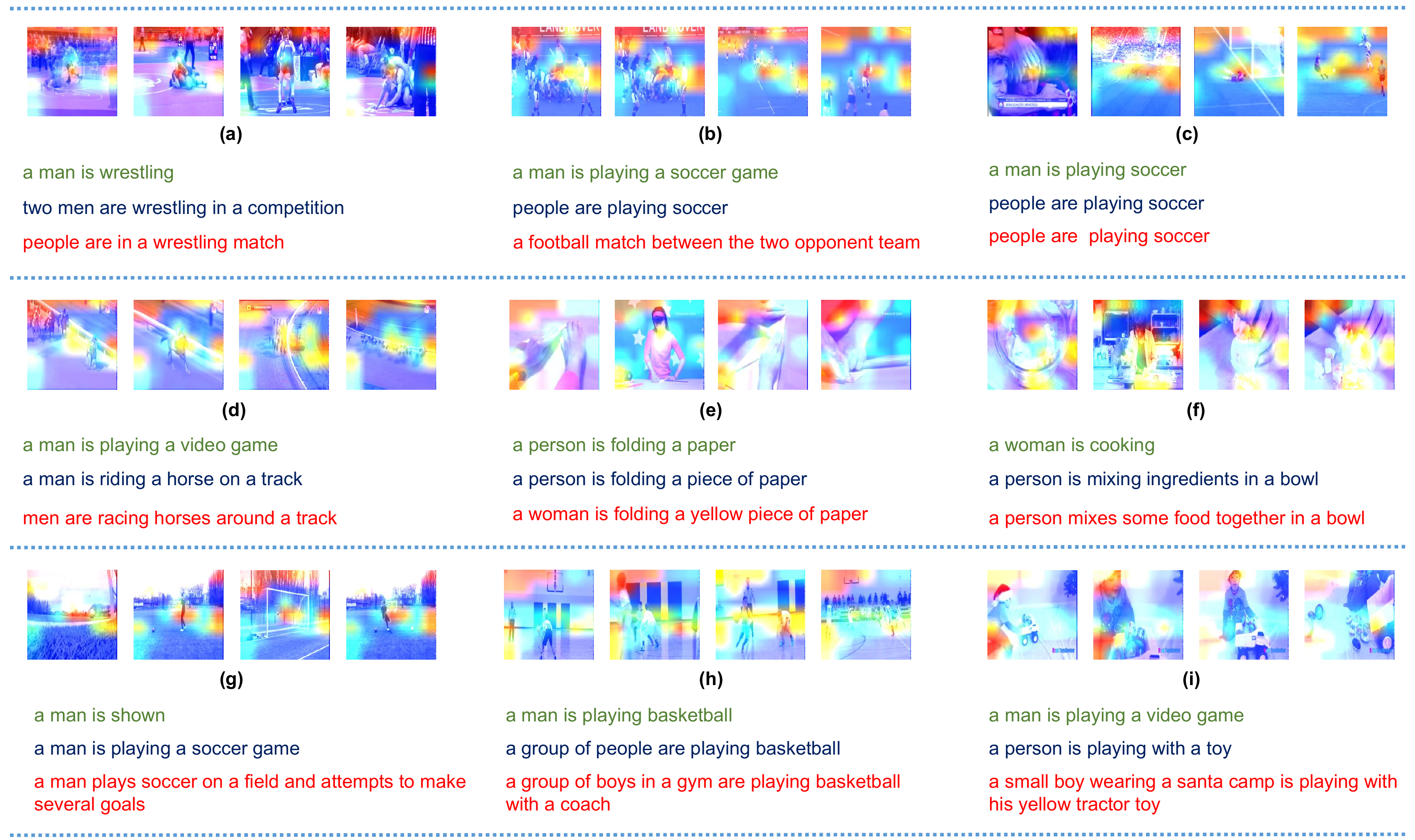}
\caption{Example results of our model on the MSR-VTT dataset. The generated captions and the ground truth are shown. The corresponding spatial attention maps are generated by our VCCV model and visualized as heatmaps. (Best viewed in color)}

\label{fig:Qualitative}
\end{figure*}

\subsection{Qualitative Analysis}
In Fig.~\ref{fig:Qualitative}, we present some video clips and their corresponding descriptions, both generated and reference. The green sentences are generated by the baseline method (I-frame model in Tab.~\ref{tab:model varient msrvtt} and Tab.~\ref{tab:model varient charades}), while the blue sentences are generated by VCCV. The red sentences are the reference captions labeled by human. From the results, we can see that our model generates relevant captions while attending to the salient regions of the frames. For example, in the case (a), the basic method only recognizes that \emph{a man is wrestling}, which is not specific for this scene. The RAE can further generate a more specific caption: \emph{two men are wrestling in a competition}. It can be seen that the captions generated by the RAE are more closed to the ground truth than that by the baseline method.


\section{Conclusion}
In this paper, we propose a deep model for video captioning directly on compressed videos. The proposed video captioning method spots salient regions of I-frames under the assistance of the spatial information in residuals frames. In addition, a gate mechanism named temporal gate module can efficiently alleviate the disturbance caused by some noisy residuals frames and retain the most essential features for caption generation. Extensive experiments conducted on MSR-VTT and Charades show that our method achieves results that rival state-of-the-art methods.

\bibliographystyle{IEEEtran}
\bibliography{IEEEfull}


\end{document}